\documentclass[runningheads]{llncs}
\usepackage{graphicx}
\usepackage{booktabs}
\usepackage{float}
\usepackage{amsmath}
\usepackage[T1]{fontenc}
\usepackage{adjustbox}
\usepackage{placeins}
\usepackage{booktabs}
\usepackage{graphicx}
\usepackage{hyperref}
\usepackage{amssymb}
\begin{document}
\title{Benchmarking Synthetic Time Series Generation Methods for Privacy-Preserving Forecasting}
\titlerunning{Privacy-Preserving Synthetic Time Series Generation}
%
    \author{Luis~Amorim\inst{2}, Vitor~Cerqueira\inst{6}, Moisés~Santos\inst{2,3},
    Carlos~Soares\inst{2,3,4},
    Paulo~J.~Azevedo\inst{1,5}}
\authorrunning{L. Amorim et al.}
    
    \institute{Department of Informatics, University of Minho, Braga, Portugal\\ 
    \and
    Faculdade de Engenharia da Universidade do Porto, Porto, Portugal \and
    Laboratory for Artificial Intelligence and Computer Science (LIACC), Portugal \and
    Fraunhofer Portugal AICOS, Portugal\\
    \and
    HASLab - INESCTEC\\ 
    \and
    University of Coimbra, Coimbra, Portugal\\ 
    }
\maketitle              
\begin{abstract}
Time series forecasting in privacy-sensitive domains often requires training models on released data rather than original observations. Synthetic time series generation has been developed primarily for data augmentation, where generated series supplement the original training set. How well these methods perform when fully replacing the original data--and how much privacy risk the released series carry--remains underexplored. We address this gap through a benchmark evaluating synthetic generation methods and noise-based anonymization baselines under a Train on Synthetic, Test on Real (TSTR) protocol. We jointly assess forecasting performance and distance-based empirical privacy risk across seven datasets, characterizing the trade-off between these objectives. We also introduce Grasynda-P, a privacy-motivated extension of the graph-based generator Grasynda, incorporating matrix ensembling and kernel density estimation. 

Our results show that: (1) no generation method fully substitutes for original training data; (2) noise-based anonymization yields the strongest privacy but the worst forecasting performance; (3) simple transformation-based generators outperform deep generative models for forecasting in this setting; and (4) Grasynda-P lies on the Pareto frontier, achieving competitive forecasting with stronger privacy separation than other generators.
This benchmark establishes a reference point for evaluating and developing new privacy-aware synthetic time series generation methods.

\keywords{Time series  \and Forecasting \and Data Privacy \and Synthetic Data \and Benchmark}
\end{abstract}
\section{Introduction}
Time series forecasting is essential in domains such as healthcare, energy, and finance~\cite{hyndman2014forecasting}. In many of these applications, access to original training data is often restricted due to privacy concerns~\cite{lpafpa}, yet accurate forecasting remains an operational requirement. Forecasting models must therefore be trained on released data: either synthetic series or privacy-protected versions of the original dataset.

Most synthetic time series generation methods have primarily been developed and evaluated as a data augmentation tool, where generated series expand the original training set to improve forecasting performance~\cite{BANDARA,victor2024enhancing}. This approach falls short when the original data cannot be used for training. How well synthetic time series generation methods perform when their output is the sole training source, and how much information about the original series they reveal, remains underexplored. 

Privacy-oriented perturbation methods such as LPA and FPA~\cite{lpafpa} approach the problem from a different angle, protecting data through noise injection or frequency-domain perturbation. However, they often remove temporal correlations~\cite{lpafpa,cao2017quantifying} in the process, which are, in principle, critical for forecasting. Whether these methods remain useful for downstream forecasting is unclear.

This paper addresses these gaps through a systematic benchmark. We evaluate synthetic generation methods and noise-based anonymization baselines under a Train on Synthetic, Test on Real (TSTR) protocol, jointly assessing both forecasting performance and empirical privacy risk across seven datasets. The central research questions in our work are: how do synthetic time series generation methods compare when used as the sole training source in forecasting problems? How private are the released series relative to the original? And does any approach achieve a favorable trade-off between forecasting utility and privacy?

Among the methods we evaluate, we include Grasynda~\cite{amorim2025grasynda}, a recently proposed graph-based generator that constructs synthetic series from transition probability matrices over discrete state spaces. Unlike direct transformation approaches, graph-based generation produces sequences driven by transition dynamics, which need not preserve point-wise similarity to the training data. We extend \texttt{Grasynda} with two privacy-motivated modifications: (1) matrix ensembling, which reduces the influence of individual training series on generation, and (2) kernel density estimation for value reconstruction, which smooths the mapping from discrete states back to continuous values. We refer to the extended method as \texttt{Grasynda-P}.

Experiments show that no generation method fully substitutes for original training data under TSTR. Simple transformation-based generators generally achieve stronger forecasting performance than the evaluated deep generative models in this setting. Noise-based anonymization methods yield the strongest empirical privacy scores but perform consistently worst for forecasting, highlighting that the two objectives are difficult to satisfy simultaneously. Among the synthetic generators, Grasynda-P lies on the Pareto frontier of the privacy--forecasting trade-off, combining competitive forecasting performance with stronger empirical privacy separation than competing methods.
This paper makes two key contributions: 
\begin{itemize}
    \item A benchmark of synthetic time series generation methods and noise-based anonymization baselines under a TSTR protocol, assessing forecasting performance and distance-based empirical privacy risk across seven datasets. This includes a characterization of the privacy--forecasting trade-off, identifying where each method positions on both dimensions.
    \item A privacy-motivated extension of Grasynda~\cite{amorim2025grasynda} that improves empirical privacy without compromising forecasting utility.
\end{itemize}

By providing a systematic comparison under realistic privacy constraints, this benchmark establishes a reference point for evaluating future synthetic generation methods in time series forecasting problems. The experiments are available in the following repository: \url{https://github.com/Amorim009/Grasynda}.

\section{Background}

\subsection{Privacy-Sensitive Forecasting}

A univariate time series is a time-ordered sequence of observations $Y = \{y_1, y_2, \ldots, y_T\}$, where $y_t \in \mathbb{R}$ is the observation at time $t$ and $T$ is the series length. Forecasting consists of predicting future values $y_{T+1}, \ldots, y_{T+h}$ given a window of past observations $y_{T-p+1}, \ldots, y_T$, where $h$ is the forecasting horizon and $p$ is the input window length (number of lags). Most forecasting problems involve collections $\mathcal{D} = \{Y^{(1)}, \ldots, Y^{(M)}\}$ of $M$ series~\cite{hyndman2014forecasting}. Global forecasting models learn a single set of parameters across all series in the collection, enabling cross-series learning and parameter sharing~\cite{januschowski2020criteria}.

In privacy-sensitive domains, using the original time series for model training may be infeasible. We consider a setting where a release method $g$ receives a private training dataset $\mathcal{D}$ and produces a released dataset:
\begin{equation}
    \tilde{\mathcal{D}} = g(\mathcal{D}) =
    \{\tilde{Y}^{(1)}, \ldots, \tilde{Y}^{(M)}\}.
\end{equation}
For synthetic generation methods, $\tilde{\mathcal{D}}$ consists of generated time series. For perturbation or anonymization methods, $\tilde{\mathcal{D}}$ consists of protected versions of the original series. 
We adopt the TSTR protocol: the forecasting model is trained exclusively on the released dataset $\tilde{\mathcal{D}}$ and evaluated on held-out observations from the original series. This protocol enables us to quantify how well the released data preserve the predictive structure needed for accurate forecasting on real data and how well $g$ preserves the privacy of original observations.

\subsection{Synthetic Time Series Generation} \label{sec:synth_methods}

Synthetic time series generation methods produce artificial data that can be used for model training. These methods have been developed primarily for data augmentation, though they may also serve as training data replacements in privacy-sensitive settings. We organize them into three categories based on their generation mechanism: transformation methods, pattern mixing methods, and deep generative models. 

Transformation methods modify individual series to create variations. Jittering adds random noise to observations, while scaling multiplies values by random factors~\cite{cerqueira2025online}. Time warping reshapes the temporal axis using cubic splines, and magnitude warping applies similar reshaping directly to values~\cite{timemagwarp}. These approaches are computationally efficient but produce series that remain structurally close to the originals~\cite{semenoglou2023data}.

Pattern mixing methods combine information across multiple series. TSMixup \cite{ansari2024chronos} blends randomly sampled segments from different series using weighted averages. Chronos \cite{ansari2024chronos}, Amazon's foundation model for forecasting, was pre-trained using synthetic data from TSMixup. DBA \cite{forestier2017generating} averages series through DTW alignment. SeasonalMBB \cite{bergmeir2016bagging} resamples blocks from decomposed series to preserve seasonal patterns. 

Deep generative models learn distributions from training data to produce synthetic sequences. TimeVAE~\cite{desai2021timevae} uses variational autoencoders with interpretable temporal components. As for diffusion models: TSDiff~\cite{TSDiff} introduces an unconditionally trained diffusion model with self-guidance at inference time. These methods can capture complex temporal dependencies but require substantial computational resources and careful hyperparameter tuning, and may be prone to overfitting and mode collapse.

\subsection{Privacy in Time Series}

Differential privacy~\cite{dwork2006calibrating,cao2017quantifying} provides a formal privacy guarantee parameterized by $\varepsilon$. A randomized algorithm $\mathcal{M}$ satisfies $\varepsilon$-differential privacy if, for any two neighboring datasets $D$ and $D'$ and any set of possible outputs $S \subseteq Range({M})$,
\[
\Pr[\mathcal{M}(D) \in S] \leq e^{\varepsilon} \Pr[\mathcal{M}(D') \in S].
\]
The parameter $\varepsilon$, called the privacy budget, represents the degree of privacy offered. A lower value of $\varepsilon$ implies a stronger privacy guarantee and typically requires larger perturbation noise. A commonly used method to achieve $\varepsilon$-DP is the Laplace mechanism, which adds random noise drawn from a Laplace distribution calibrated to the sensitivity of the released quantity~\cite{dwork2006calibrating}.

For time-series data, privacy protection is particularly challenging because observations are temporally correlated. Mechanisms calibrated according to standard differential privacy assumptions may experience additional privacy leakage when adversaries exploit such correlations~\cite{cao2017quantifying}.

Different generative models for privacy-preserving synthetic data generation have been proposed and benchmarked~\cite{yoon2018pategan,tao2022benchmarking}, but not for time series generation or downstream forecasting tasks.

The Laplace Perturbation Algorithm (LPA)~\cite{lpafpa} applies this directly to time-series observations, adding independent Laplace noise at each time step. The Fourier Perturbation Algorithm (FPA)~\cite{lpafpa} operates in the frequency domain, perturbing low-frequency Fourier coefficients to reduce the noise required for longer sequences. These methods are not designed to produce released data suitable for training forecasting models. LPA injection can alter correlation structure and local temporal dependencies, which may reduce forecasting utility, while FPA introduces reconstruction errors ~\cite{lpafpa}.


\section{Graph-Based Privacy-Aware Generation}
\label{sec:graph_generation}

We use Grasynda~\cite{amorim2025grasynda}, a transition-graph generator recently introduced for data augmentation, and evaluate it under TSTR where generated series fully replace the original training data. Graph-based generation encodes temporal dynamics through transition probability matrices over discrete state spaces, separating the generation mechanism from the original value sequences. We propose two privacy-motivated modifications: matrix ensembling, which alters the transition matrix used for generation, and continuous state-value sampling, which changes how sampled states are converted back to values.  

\subsection{Base Generator}\label{sec:base_model}

Given a univariate time series $Y = \{y_1, \ldots, y_T\}$, Grasynda maps each observation $y_t$ to a discrete state $s_t \in \{1, \ldots, k\}$ via quantile-based discretization, where $k$ is a hyperparameter controlling the granularity of the state space. For each state $j$, let $R_j = \{ y_t \in Y \mid s_t = j \}$ denote the set of original values assigned to that state.

The state sequence defines a directed transition graph where nodes are states and edges are observed transitions. The transition matrix $P = [p_{ij}]$ encodes the probability of moving from state $i$ to state $j$:
\begin{equation}
    p_{ij} =
    \frac{\mathrm{count}(i \rightarrow j)}
    {\sum_{j'=1}^{k} \mathrm{count}(i \rightarrow j')}.
\end{equation}
Synthetic sequences of the same length as the original are generated through three steps: (1) the initial state $\hat{s}_1$ is sampled from the stationary distribution of $P$; (2) subsequent states are sampled via transitions, where $\hat{s}_{t+1} \sim P(\hat{s}_t)$; and (3) each state $\hat{s}_t$ is mapped back to a continuous value by sampling uniformly from $R_{\hat{s}_t}$. This final step may reproduce exact values from the original series.

\begin{figure}[t]
    \centering

    \begin{minipage}[t]{0.37\linewidth}
        \centering
        \includegraphics[width=\linewidth]{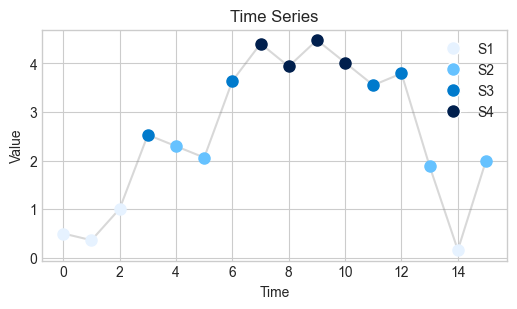}
        \vspace{1mm}
        \small (a)
    \end{minipage}
    \hfill
    \begin{minipage}[t]{0.33\linewidth}
        \centering
        \includegraphics[width=\linewidth]{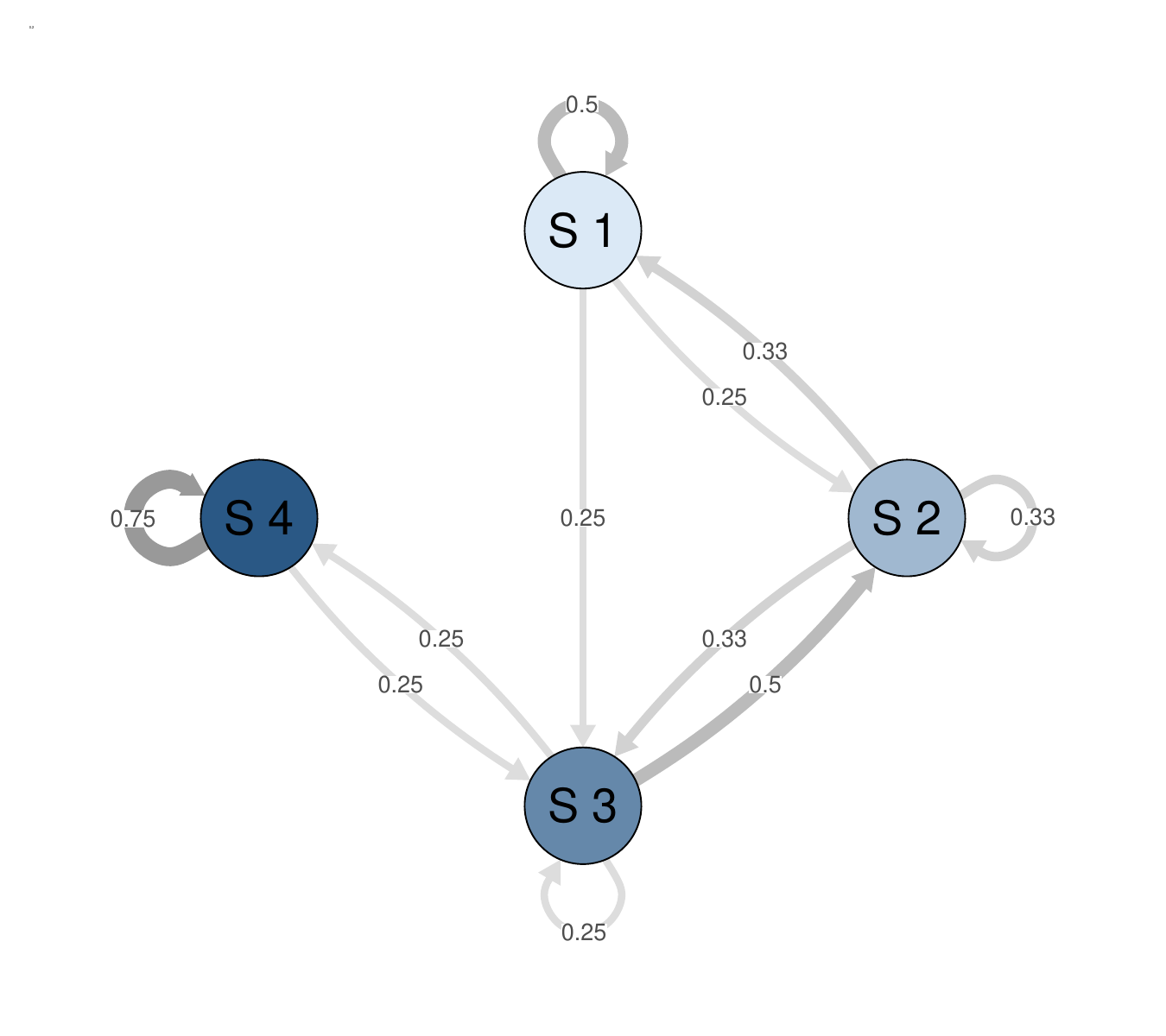}
        \vspace{1mm}
        \small (b)
    \end{minipage}
    \hfill
    \begin{minipage}[t]{0.24\linewidth}
        \centering
        \includegraphics[width=\linewidth]{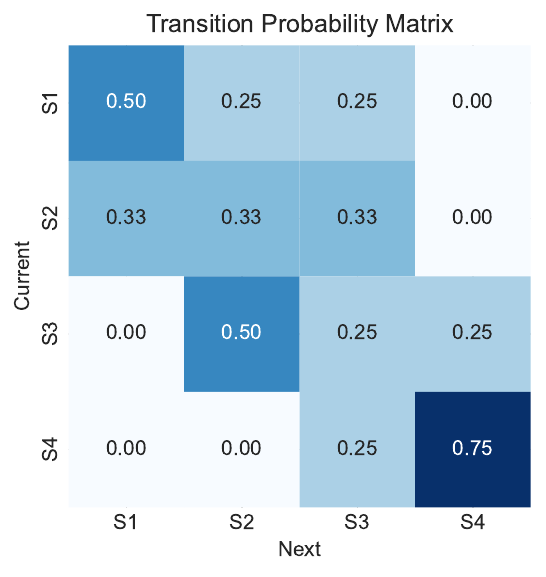}
        \vspace{1mm}
        \small (c)
    \end{minipage}

    \caption{Grasynda pipeline: (a) original time series, (b) quantile graph representation, (c) transition matrix used for generation.}
    \label{fig:exampleONE}
\end{figure}

\subsection{Matrix Ensembling}

The base method computes a transition matrix from a single series, meaning the generated sequence depends entirely on one training record. We replace this with an ensemble approach that reduces the influence of any individual series. Given a target series $Y^{(a)}$ with transition matrix $P^{(a)}$, we identify the $r$ most similar matrices from the training set using Frobenius distance:
\begin{equation}
    d_{ab} = \| P^{(a)} - P^{(b)} \|_F
    = \sqrt{\sum_{m=1}^{k} \sum_{n=1}^{k}
    \left( p_{mn}^{(a)} - p_{mn}^{(b)} \right)^2 }.
\end{equation}
The ensemble matrix averages $P^{(a)}$ with its $r$ nearest neighbors:
\begin{equation}
    \bar{P}^{(a)} =
    \frac{1}{r+1}
    \left(
    P^{(a)} + \sum_{b \in \mathcal{N}_a} P^{(b)}
    \right),
\end{equation}
where $\mathcal{N}_a$ indexes the $r$ nearest matrices excluding $a$. From a privacy perspective, ensembling dilutes the contribution of any single training series to the generation process, making it harder to infer properties of individual records from the synthetic output. As a secondary benefit, ensembling smooths extreme transition probabilities and stabilizes estimation when individual series are short.

\subsection{Continuous State-Value Sampling}

The uniform sampling step in the base method can reproduce exact values from the original series, creating a direct link between synthetic and real observations. We replace this with kernel density estimation~\cite{silverman2018density}, which generates values from a smoothed distribution rather than the discrete set of originals. For each state $j$ with observed values $R_j = \{r_1, \ldots, r_{m_j}\}$, we fit a Gaussian KDE:
\begin{equation}
    f_j(x) =
    \frac{1}{m_j b_j}
    \sum_{i=1}^{m_j}
    K\!\left(\frac{x-r_i}{b_j}\right),
\end{equation}
where $K$ is the Gaussian kernel and $b_j$ is the bandwidth selected via Scott's rule~\cite{scott1992multivariate}. Synthetic values are sampled as $\hat{y}_t \sim f_{\hat{s}_t}$, producing continuous outputs that are not constrained to original observations. This modification prevents the reproduction of exact values, increasing the distance between synthetic and original series.

The two modifications lead to \texttt{Grasynda-P}, a generator that addresses privacy at both the structural and value levels. Matrix ensembling reduces dependence on individual training series, while KDE sampling avoids exact reproduction of original values, creating a stronger separation between synthetic and original series. The hyperparameters $k$ (number of states) and $r$ (number of neighbors for ensembling) are tuned via random search as described in the next Section.

\section{Experimental Design}

We benchmark synthetic time series generation methods and noise-based anonymization baselines under a TSTR protocol, where released data fully replace the original training series. We address four research questions:

\begin{itemize}
    \item \textbf{RQ1.} How do synthetic generation methods and noise-based anonymization baselines compare in forecasting accuracy under a TSTR protocol?

    \item \textbf{RQ2.} How do these methods compare in terms of distance-based empirical privacy risk?

    \item \textbf{RQ3.} How strong is the trade-off between forecasting utility and privacy separation, and which methods lie on the Pareto frontier?

    \item \textbf{RQ4.} Does \texttt{Grasynda-P} improve empirical privacy over \texttt{Grasynda} without compromising forecasting performance?
\end{itemize}

\subsection{Datasets}

We use seven publicly available datasets from four forecasting competitions: M1~\cite{m1comp}, M3~\cite{m3competition}, Tourism~\cite{athanasopoulos2011tourism}, and NN3~\cite{NN3}, spanning industry, demography, and economics domains. Table~\ref{tab:data} summarizes their characteristics. The forecasting horizon $h$ is 12 for monthly and 8 for quarterly series; the input window is 24 and 8 respectively, corresponding to two seasonal cycles.

\begin{table}[bt]
\centering
\caption{Summary of the datasets: number of time series, observations, seasonal period, and forecasting horizon ($h$).}
\label{tab:data}
\begin{tabular}{lrrrr}
\toprule
Dataset & \# Series & \# Obs. & Period & $h$ \\
\midrule
M1-M  & 617  & 44,892  & 12 & 12 \\
M1-Q  & 203  & 8,320   & 4  & 8  \\
M3-M  & 1428 & 167,562 & 12 & 12 \\
M3-Q  & 756  & 37,004  & 4  & 8  \\
T-M   & 366  & 109,280 & 12 & 12 \\
T-Q   & 427  & 42,544  & 4  & 8  \\
NN3   & 111  & 10,041  & 12 & 12 \\
\midrule
Total & 3908 & 419,643 & -- & -- \\
\bottomrule
\end{tabular}
\end{table}

\subsection{Methods}

\paragraph{Synthetic generators.} We evaluate \texttt{Scaling}~\cite{semenoglou2023data}, \texttt{Jitter}~\cite{semenoglou2023data}, \texttt{M-Warp}~\cite{timemagwarp}, \texttt{T-Warp}~\cite{timemagwarp}, \texttt{DBA}~\cite{forestier2017generating}, \texttt{SeasonalMBB}~\cite{bergmeir2016bagging}, \texttt{TSMixup}~\cite{ansari2024chronos}, \texttt{TimeVAE}~\cite{desai2021timevae}, and \texttt{TSDiff}~\cite{TSDiff}, alongside both \texttt{Grasynda} and \texttt{Grasynda-P}. This covers classical transformation-based methods and deep generative models. 

\paragraph{Baselines.}
We include \texttt{LPA} and \texttt{FPA} to 
contextualize the privacy--forecasting trade-off 
against methods designed for disclosure protection. 
Following~\cite{lpafpa}, we set $\varepsilon = 1$. 
\texttt{LPA} adds Laplace noise with scale $ns/\varepsilon$ 
($n$: series length, $s$: per-timestamp sensitivity). 
\texttt{FPA} applies an orthonormal FFT, retains 
$k=30$ low-frequency coefficients, and perturbs 
coordinates with scale $\sqrt{2k}\sqrt{n}\,s/\varepsilon$. 
Since clipping bounds are data-derived, these are 
DP-inspired empirical baselines rather than formal 
differentially private releases.

Training on \texttt{Original} data provides a non-private forecasting upper bound. We also include a set of series of random noise serves as a sanity check for privacy metrics.
Each method produces one released series per original training series. Synthetic generator parameters are optimized via 40 iterations of random search; noise-based baselines use $\varepsilon = 1$ following~\cite{lpafpa} and \cite{cao2017quantifying}.

\subsection{Forecasting Protocol}

\begin{figure}[tb]
    \centering
    \includegraphics[width=.6\linewidth]{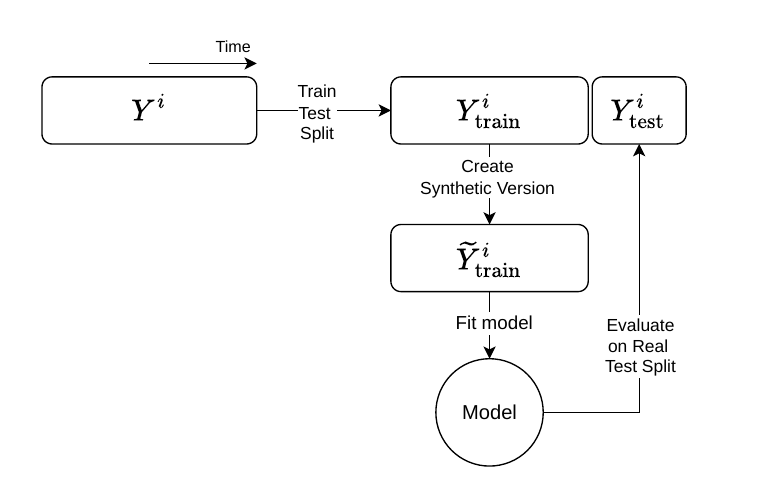}
    \caption{TSTR protocol for evaluating synthetic time series generators for an example time series $Y^{(i)}$.}
    \label{fig:protocol}
\end{figure}

For each dataset, the final $h$ observations of each series are held out for testing, leading to the test set $\mathcal{D}_{test}$. Given training set $\mathcal{D}_{train} = \{Y_{train}^{(1)}, \ldots, Y_{train}^{(M)}\}$, each method produces a released dataset $\tilde{\mathcal{D}}_{train} = \{ \tilde{Y}_{train}^{(1)}, \ldots, \tilde{Y}_{train}^{(M)} \}$, where $\tilde{Y}_{train}^{(i)}$ is either synthetic or a perturbed version of $Y_{train}^{(i)}$. Forecasting models are trained on $\tilde{\mathcal{D}}_{train}$ and evaluated on the real held-out observations ($\mathcal{D}_{test}$) using Mean Absolute Scaled Error (MASE)~\cite{hyndman2006another}, where lower values indicate better performance. We use NHITS~\cite{nhits} as the forecasting model, implemented in \texttt{neuralforecast}. This method have shown to perform competitively for forecasting in benchmark datasets~\cite{zeng2023transformers}. The protocol is illustrated in Figure~\ref{fig:protocol}.

\subsection{Privacy protocol}

We measure empirical privacy using Distance to Closest Record (DCR) and Nearest Neighbour Distance Ratio (NNDR). For each released series $\tilde{Y}_{train}^{(j)}$, let $d_{j,(1)} \leq d_{j,(2)}$ denote the Euclidean distances to its nearest and second-nearest original series, computed on z-score normalized series:
\begin{equation}
    \mathrm{DCR}_j = d_{j,(1)}, \qquad
    \mathrm{NNDR}_j = \frac{d_{j,(1)}}{d_{j,(2)}}.
\end{equation}
Higher DCR indicates greater distance from the closest original record; higher NNDR indicates more uniform distances to multiple originals, reducing single-record dominance. We report means over all released series. The differences between \texttt{Grasynda-P} and \texttt{Grasynda} are tested for significance using the Wilcoxon signed-rank test.

We compute separate ranks for forecasting and privacy within each dataset. Spearman correlation between average forecasting and privacy ranks quantifies the strength of the trade-off across methods. We further characterize favorable positions using Pareto dominance: a method is Pareto-dominated if another method achieves better or equal performance on both dimensions and strictly better on at least one.

\section{Results}

\subsection{Forecasting Evaluation}
\begin{table}[t]
\centering
\caption{NHITS architecture: TSTR forecasting performance and average rank. Best values are in bold and second-best values are underlined.}
\label{tab:nhits_tstr_final}
\resizebox{.8\textwidth}{!}{%
\begin{tabular}{l|ccccccc|c}
\toprule
\textbf{Method} & \textbf{M3-M} & \textbf{M3-Q} & \textbf{T-M} & \textbf{T-Q} & \textbf{M1-M} & \textbf{M1-Q} & \textbf{NN3-M} & \textbf{Avg Rank} \\ \midrule
Original           & \textbf{0.718} & \underline{0.832} & 1.189 & 1.229 & \underline{0.984} & \textbf{1.317} & \underline{1.066} & \textbf{2.21} \\
Jitter             & 0.726 & 0.833 & 1.205 & 1.229 & \textbf{0.982} & 1.332 & 1.094 & \underline{3.64} \\
Scaling            & 0.772 & 0.972 & \underline{1.186} & \textbf{1.218} & 1.018 & 1.393 & \textbf{1.047} & 5.29 \\
SeasMBB            & 0.734 & 0.854 & 1.198 & 1.250 & 1.061 & 1.364 & 1.097 & 5.71 \\
Grasynda-P         & 0.736 & 0.871 & 1.200 & 1.228 & 1.067 & 1.357 & 1.107 & 6.00 \\
TSMix              & 0.750 & 0.855 & \textbf{1.182} & 1.237 & 1.106 & \underline{1.328} & 1.241 & 6.07 \\
MagWarp            & \underline{0.722} & 0.896 & 1.199 & \underline{1.223} & 1.106 & 1.353 & 1.183 & 6.07 \\
Grasynda           & 0.725 & 0.874 & 1.214 & 1.234 & 1.165 & 1.338 & 1.095 & 6.29 \\
DBA                & 0.738 & \textbf{0.822} & 1.222 & 1.236 & 1.063 & 1.351 & 1.250 & 6.43 \\
TimeWarp           & 0.846 & 0.872 & 2.241 & 1.295 & 1.090 & 1.358 & 1.118 & 9.00 \\
TimeVAE            & 0.809 & 0.861 & 1.287 & 1.557 & 1.277 & 1.638 & 1.106 & 9.71 \\
TSDiff             & 0.906 & 1.114 & 2.914 & 2.982 & 1.201 & 1.471 & 1.496 & 12.00 \\
FPA ($\varepsilon$=1.0) & 1.245 & 1.401 & 3.076 & 2.712 & 1.390 & 1.622 & 1.733 & 13.14 \\
LPA ($\varepsilon$=1.0) & 1.224 & 1.398 & 3.089 & 3.625 & 1.405 & 1.670 & 1.381 & 13.43 \\
\bottomrule
\end{tabular}%
}
\end{table}

Table~\ref{tab:nhits_tstr_final} reports TSTR forecasting 
performance under the NHITS architecture. As expected, 
training on original data (\texttt{Original}) achieves the 
best average rank, confirming that no generation method 
fully substitutes for original training data across all 
datasets. In general, methods that produce series closer 
to the original tend to perform better, with simple 
transformation methods such as \texttt{Jitter} and 
\texttt{Scaling} outperforming more complex deep generative 
approaches such as \texttt{TimeVAE} and \texttt{TSDiff}.

\texttt{Grasynda-P} ranks competitively among generation methods, ahead of \texttt{TSMixup},  \texttt{MagWarp}, and \texttt{DBA}. This modified version achieves a slightly better average rank than the original Grasynda. A Wilcoxon signed-rank test on per-series MASE differences reveals no statistically significant difference in forecasting performance between the two versions.
Noise-based anonymization methods perform consistently worst across all datasets, with \texttt{LPA} and \texttt{FPA} at all noise levels occupying the bottom ranks and producing particularly large MASE values on several datasets, confirming that perturbation-based methods are not suitable as training data substitutes for forecasting.

\subsection{Privacy Evaluation  }

\begin{table}[tb]
\centering
\caption{Privacy evaluation with DCR and NNDR per dataset and average rank. Higher values indicate greater empirical privacy separation. Best values are in bold and second-best underlined. $^{*}$ in Grasynda-P denotes a statistically significant improvement over Grasynda (Wilcoxon signed-rank test, $p < 0.05$).}
\label{tab:dcr_nndr_privacy}
\resizebox{\textwidth}{!}{%
\begin{tabular}{l|cc|cc|cc|cc|cc|cc|cc|c|c}
\toprule
& \multicolumn{2}{c|}{\textbf{M3-M}} & \multicolumn{2}{c|}{\textbf{M3-Q}} & \multicolumn{2}{c|}{\textbf{T-M}} & \multicolumn{2}{c|}{\textbf{T-Q}} & \multicolumn{2}{c|}{\textbf{M1-M}} & \multicolumn{2}{c|}{\textbf{M1-Q}} & \multicolumn{2}{c|}{\textbf{NN3-M}} & \textbf{DCR} & \textbf{NNDR} \\
\textbf{Method} & DCR & NNDR & DCR & NNDR & DCR & NNDR & DCR & NNDR & DCR & NNDR & DCR & NNDR & DCR & NNDR & \textbf{Rank} & \textbf{Rank} \\
\midrule
Original & 0.000 & 0.001 & 0.000 & 0.007 & 0.000 & 0.000 & 0.000 & 0.000 & 0.000 & 0.000 & 0.000 & 0.000 & 0.000 & 0.000 & - & - \\
Random Noise & 104.35 & 0.968 & 14.44 & 0.977 & 76.27 & 0.919 & 52.27 & 0.863 & 37.18 & 0.892 & 11.12 & 0.932 & 69.14 & 0.957 & - & - \\
LPA ($\varepsilon$=1.0) & \textbf{14.561} & \textbf{0.989} & \underline{9.387} & \textbf{0.979} & \textbf{14.452} & \textbf{0.985} & \textbf{2.517} & \textbf{0.972} & \underline{12.604} & \textbf{0.989} & \textbf{9.800} & \textbf{0.985} & \underline{17.631} & \textbf{0.984} & \textbf{1.43} & \textbf{1.14} \\
FPA ($\varepsilon$=1.0) & \underline{14.371} & \underline{0.988} & \textbf{9.422} & \underline{0.978} & \underline{13.886} & \underline{0.975} & \underline{2.497} & \underline{0.971} & \textbf{12.631} & \textbf{0.989} & \underline{9.768} & \underline{0.984} & \textbf{17.756} & \textbf{0.984} & \underline{1.57} & \underline{1.86} \\
Grasynda-P & 2.461$^{*}$ & 0.898 & 1.779$^{*}$ & 0.861 & 0.991$^{*}$ & 0.885$^{*}$ & 0.175$^{*}$ & 0.863$^{*}$ & 0.381$^{*}$ & 0.791$^{*}$ & 0.218$^{*}$ & 0.730$^{*}$ & 5.079$^{*}$ & 0.827$^{*}$ & 3.57 & 5.57 \\
TSDiff & 1.606 & 0.952 & 0.859 & 0.930 & 0.490 & 0.927 & 0.107 & 0.908 & 0.299 & 0.891 & 0.107 & 0.811 & 2.543 & \underline{0.928} & 6.71 & 3.71 \\
Grasynda & 1.753 & 0.889 & 0.791 & 0.821 & 0.599 & 0.779 & 0.136 & 0.787 & 0.256 & 0.714 & 0.113 & 0.652 & 2.484 & 0.757 & 5.86 & 7.00 \\
TimeVAE & 1.248 & 0.818 & 0.670 & 0.801 & 1.637 & 0.955 & 0.286 & 0.961 & 0.443 & \underline{0.960} & 0.173 & 0.938 & 1.454 & 0.838 & 6.00 & 4.57 \\
DBA & 1.633 & 0.723 & 0.764 & 0.661 & 0.241 & 0.360 & 0.119 & 0.355 & 0.494 & 0.659 & 0.068 & 0.374 & 1.824 & 0.589 & 7.14 & 9.71 \\
TSMix & 1.235 & 0.924 & 0.636 & 0.753 & 0.547 & 0.906 & 0.124 & 0.868 & 0.233 & 0.816 & 0.133 & 0.647 & 1.757 & 0.770 & 7.86 & 6.00 \\
TimeWarp & 1.826 & 0.902 & 0.728 & 0.735 & 0.986 & 0.940 & 0.086 & 0.767 & 0.187 & 0.639 & 0.016 & 0.286 & 1.308 & 0.481 & 8.29 & 8.29 \\
SeasMBB & 1.655 & 0.720 & 0.579 & 0.578 & 0.275 & 0.497 & 0.037 & 0.429 & 0.127 & 0.459 & 0.051 & 0.384 & 1.913 & 0.629 & 9.57 & 10.00 \\
MagWarp & 0.468 & 0.400 & 1.275 & 0.852 & 0.149 & 0.196 & 0.038 & 0.214 & 0.130 & 0.514 & 0.054 & 0.584 & 0.632 & 0.298 & 10.00 & 9.86 \\
Scaling & 1.102 & 0.599 & 1.465 & 0.856 & 0.094 & 0.106 & 0.023 & 0.177 & 0.134 & 0.486 & 0.056 & 0.573 & 0.197 & 0.103 & 10.00 & 10.29 \\
Jitter & 0.028 & 0.034 & 0.248 & 0.371 & 0.061 & 0.070 & 0.019 & 0.149 & 0.040 & 0.168 & 0.004 & 0.111 & 0.180 & 0.063 & 13.00 & 13.00 \\
\bottomrule
\end{tabular}%
}
\end{table}

Table~\ref{tab:dcr_nndr_privacy}  reports empirical privacy results across all methods. As expected, noise-based anonymization methods achieve the highest privacy separation, given that they directly perturb the original series. Among synthetic generation methods, \texttt{Grasynda-P}, \texttt{TSDiff} and \texttt{TimeVAE} are the most competitive. \texttt{Grasynda-P} ranks first among generators on DCR; and \texttt{TSDiff} ranks first on NNDR, followed by \texttt{TimeVAE} and \texttt{Grasynda-P}. Transformation-based methods such as \texttt{Jitter} and \texttt{Scaling} rank consistently lowest, as these simple operations produce series that remain close to the original observations.

\texttt{Grasynda-P} achieves higher empirical privacy than 
\texttt{Grasynda} across all metrics, indicating that the proposed method changes contribute to privacy improvements. This improvement is statistically significant on DCR across all seven datasets (Wilcoxon signed-rank test, $p < 0.05$). NNDR improvements are significant on five of seven datasets, though 
Grasynda-P consistently shows higher values across all datasets.

\subsection{Privacy--Forecasting Trade-off}
\begin{figure}[tb]
    \centering
    \includegraphics[width=.8\linewidth]{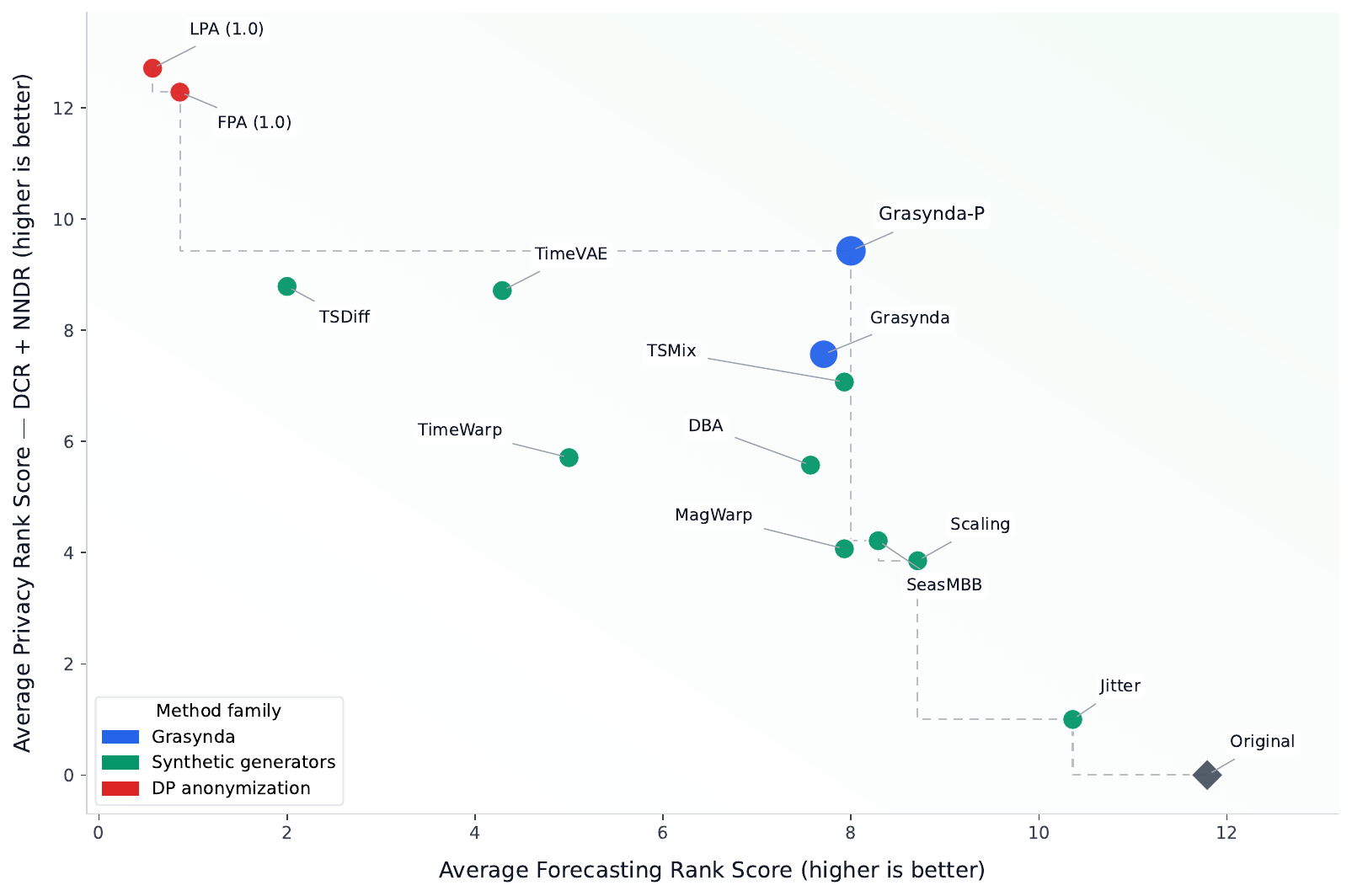}
    \caption{Average forecasting rank score vs. average empirical privacy rank score (DCR and NNDR) under the NHITS architecture, across seven benchmark datasets. Rank scores are computed as $(N - \bar{r})$, where $N$ is the number of evaluated methods and $\bar{r}$ is the average rank across datasets, so that higher values indicate better performance. The dashed staircase traces the Pareto frontier - the set of methods not dominated by any other on both objectives simultaneously.}
    \label{fig:tradeoff}
\end{figure}

Figure~\ref{fig:tradeoff} visualizes the joint position of each method across both evaluation dimensions. 

\texttt{Original}, included as a non-private forecasting reference, sits in the lower-right region. This sanity check behaves as expected: unprotected real data yields the strongest forecasting utility but has no empirical privacy. 

The distribution of methods across the plot reveals a systematic  trade-off: methods with stronger empirical privacy separation tend to exhibit weaker forecasting utility, and conversely, methods with competitive forecasting performance tend to carry higher empirical similarity to the original series. A Spearman rank correlation of $\rho = -0.776$ ($p = 0.002$, $N = 13$)  between average forecasting rank and average privacy rank across all evaluated methods confirms this trade-off. The correlation remains significant when restricting the analysis to synthetic generators only ($\rho = -0.638$, $p = 0.035$, $N = 11$), indicating the trade-off is not being driven solely by the positions of noise-based methods.

Noise-based anonymization methods occupy the upper-left region of the trade-off space, achieving high privacy separation but low forecasting utility, with the two variants clustering closely together, suggesting this trade-off is structural rather than sensitive to algorithm or noise level choice.

Among synthetic generators, simple transformation-based methods sit in the lower-right region, achieving strong forecasting performance but low privacy separation, reflecting that their outputs remain close to the original series by construction. Pattern mixing and deep generative methods occupy a more central position: \texttt{TSMix}, \texttt{TSDiff}, and \texttt{TimeVAE}  achieve higher empirical privacy separation but fall below the leading generators on forecasting utility.

\texttt{Grasynda-P} lies on the Pareto frontier and is not dominated by any other synthetic generator: no evaluated generator simultaneously achieves better forecasting utility and better empirical privacy separation. \texttt{Grasynda} is Pareto-dominated by \texttt{Grasynda-P} on both dimensions, with the proposed modifications shifting it to the frontier without a corresponding loss in forecasting utility.

\section{Discussion}

Privacy-sensitive forecasting requires training models on released data rather than original observations, yet how well existing synthetic generation methods serve this purpose—and how much privacy protection they offer—has remained unclear. We addressed this question through an extensive benchmark under a TSTR protocol, jointly evaluating forecasting utility and distance-based empirical privacy risk.

\paragraph{Research Questions}

Regarding forecasting performance (\textbf{RQ1}), no synthetic generation method fully substitutes for original training data. Simple transformation-based methods achieve higher forecasting utility than deep generative models, likely because limited variation keeps synthetic series closer to the original distribution. Notably, \texttt{Grasynda-P} ranks competitively among the strongest generators, indicating that graph-based generation preserves temporal properties important for model generalization. Noise-based baselines perform consistently worst, suggesting that direct value perturbation disrupts the temporal structure forecasting models depend on.

Concerning empirical privacy risk (RQ2), noise-based methods yield the highest empirical separation through explicit value modification rather than any generative property. Among synthetic generators, \texttt{Grasynda-P} ranks first on DCR. Its strong separation likely stems from the generation mechanism itself: operating on transition probability matrices over discrete state spaces and sampling from continuous distributions rather than reproducing original values. \texttt{TSDiff} leads on NNDR, possibly because higher intrinsic noise produces sequences with more evenly distributed nearest-neighbour distances. Transformation-based methods rank lowest, as their simple modifications preserve proximity to original observations.

The Spearman rank correlation confirms a systematic trade-off between forecasting utility and empirical privacy separation, persistent even among synthetic generators alone (\textbf{RQ3}). Methods that preserve structural proximity tend to perform well for forecasting but offer low privacy separation, while methods with strong privacy separation introduce may introduce distributional shift that impairs forecasting utility. \texttt{Grasynda-P} lies on the Pareto frontier, indicating that graph-based generation can effectively balance this trade-off.

Finally, regarding the proposed method (\textbf{RQ4}), matrix ensembling and continuous state-value sampling via kernel density estimation improve empirical privacy with statistically significant privacy gains across all seven datasets ($p < 0.05$), without compromising forecasting performance  ($p > 0.05$). KDE sampling may avoid exact reproduction of original records, while matrix ensembling reduces dependence on individual training series. \texttt{Grasynda-P} retains competitive forecasting utility, suggesting that capturing transition structures preserves the temporal dynamics needed for downstream forecasting, even under privacy-motivated modifications. These results support the hypothesis that graph-based time series generation represents a favorable and underexplored approach for privacy-sensitive time series settings.

\paragraph{Limitations}

Several limitations should be considered when interpreting these results. First, empirical privacy is measured through distance-based metrics (DCR and NNDR), which quantify proximity between released and original series but do not provide formal differential privacy guarantees. High DCR and NNDR values indicate that released series are not close copies of training records, but offer no provable bound on adversarial inference. Connecting empirical and formal privacy perspectives remains an open direction.

Second, the benchmark covers univariate monthly and quarterly series from established forecasting competitions, representing a standard evaluation setting in the literature. The findings may not generalize directly to multivariate series, higher-frequency data, or series with different temporal characteristics.  

Third, we use a single forecasting architecture (NHITS) to ensure controlled comparison across methods. While NHITS performs competitively on these benchmarks (cf.~\cite{cerqueira2025modelradar}), the relative ranking of generation methods may vary under different forecasting models.

Finally, the two modifications defining \texttt{Grasynda-P} are evaluated jointly; isolating the individual contribution of matrix ensembling and KDE sampling to the observed privacy improvements remains for future work.

\paragraph{Final Remarks}

The benchmark conducted in this paper provides a systematic comparison of synthetic time series generation methods and noise-based anonymization baselines for privacy-sensitive forecasting under a TSTR protocol. The results reveal a consistent trade-off between forecasting utility and empirical privacy separation: methods that preserve proximity to the original data excel at forecasting but offer weak privacy guarantees, while methods with strong privacy separation often degrade forecasting performance. We believe this benchmark serves as a reference point for developing and evaluating future methods for privacy-preserving synthetic time series data generation.

Among the evaluated approaches, graph-based generation via \texttt{Grasynda-P} achieves a favorable position on the Pareto frontier, combining competitive forecasting performance with stronger privacy separation than other synthetic generators. This suggests that operating on transition structures rather than direct value transformations offers a promising design principle for privacy-aware time series synthesis.

\begin{credits}
\subsubsection{\ackname} This work was partially funded by projects AISym4Med (101095387) supported by Horizon Europe Cluster 1: Health, ConnectedHealth (n.º 46858), supported by Competitiveness and Internationalisation Operational Programme (POCI) and Lisbon Regional Operational Programme (LISBOA 2020), under the PORTUGAL 2020 Partnership Agreement, through the European Regional Development Fund (ERDF) and Agenda “Center for Responsible AI”, nr. C645008882-00000055, investment project nr. 62, financed by the Recovery and Resilience Plan (PRR) and by European Union -  NextGeneration EU, and also by FCT plurianual funding for 2020-2023 of LIACC (UIDB/00027/2020 UIDP/00027/2020); This work is partially funded by national funds through FCT – Foundation for Science and Technology, I.P., within the scope of the research unit UID/00326 - Centre for Informatics and Systems of the University of Coimbra, \url{https://doi.org/10.54499/UID/00326/2025}.
\end{credits}

%
%
        
       \FloatBarrier

        \bibliographystyle{splncs04}
        \bibliography{mybibliography}

\end{document}